\documentclass[%
 reprint,
 showkeys,
 amsmath,amssymb,
 aps,
 floatfix,
]{revtex4-1}

\usepackage{graphicx}
\usepackage{dcolumn}
\usepackage{bm}
\usepackage{hyperref}
\newcommand{\PG}{p_{\mathrm{PG}}}
\newcommand{\Po}{P_{\mathrm{Po}}}
\newcommand{\GIG}{p_{\mathrm{GIG}}}

\begin{document}

\preprint{APS/123-QED}

\title{Inverse Ising problem  in continuous time: A latent variable approach}

\author{Christian Donner}
 \altaffiliation[Also at ]{Bernstein Center for Computational Neuroscience}
 \email{christian.donner@bccn-berlin.de}
\author{Manfred Opper}
\affiliation{%
 Artificial Intelligence Group, Technische Universit\"at Berlin, Germany 
}%




\date{\today}

\begin{abstract}
We consider the inverse Ising problem, i.e. the inference of network couplings from observed spin trajectories for a model with continuous time Glauber dynamics. By introducing two sets of auxiliary latent random variables we render the likelihood into a form, which allows for simple iterative inference algorithms with analytical updates. The variables are: (1) Poisson variables to linearise an exponential term which is typical for point process likelihoods and (2) P\'olya--Gamma variables, which make the likelihood quadratic in the coupling parameters. Using the augmented likelihood, we derive an expectation--maximization (EM) algorithm to obtain the maximum likelihood estimate of network parameters. Using a third set of latent variables we extend the EM algorithm to sparse couplings via L1 regularization. Finally, we develop an efficient approximate Bayesian inference algorithm using a variational approach. We demonstrate the performance of our algorithms on data simulated from an Ising model. For data which are simulated from a more biologically plausible network with spiking neurons, we show that the Ising model captures well the low order statistics of the data and how the Ising couplings are related to the underlying synaptic structure of the simulated network.

\end{abstract}

\pacs{Valid PACS appear here}
\keywords{Ising model, Glauber dynamics, Poisson process, P\'olya--Gamma, Expectation--Maximization, L1 Regularization, Bayesian Inference, recurrent neural network}
\maketitle


\section{\label{sec:intro} Introduction}
In recent years, the inverse Ising problem, i.e. the reconstruction of couplings and external fields of an Ising model from samples of spin configurations, has attracted considerable interest in the physics community \cite{nguyen2017inverse}. This is due to the fact that Ising models play an important role for data modeling with applications to neural spike data \cite{schneidman2006weak,roudi2009ising}, protein structure determination \cite{weigt2009identification},
and gene expression analysis \cite{lezon2006using}. Much effort has been devoted to the development of  algorithms for the {\em static} inverse Ising problem. This is a nontrivial task, because statistically efficient, likelihood based methods become computationally infeasible by the intractability of the  partition function of the model. Hence one has to resort to either approximate inference methods or to other statistical estimators such as pseudo--likelihood methods \cite{bachschmid2017statistical}, or the  interaction screening algorithm \cite{vuffray2016interaction}. The situation is somewhat simpler for the dynamical inverse Ising problem, which recently attracted attention \cite{roudi2011dynamical,mezard2011exact,roudi2011mean,zeng2011network,tyrcha2013effect,keshri2013shotgun}. If one assumes a Markovian dynamics, the exact normalisation of the spin transition probabilities allows for an explicit computation of the likelihood if one has a complete set of observed data over time.
Nevertheless, the model parameters enter the likelihood in a fairly complex way, and the application of more advanced statistical approaches such as Bayesian inference again becomes a nontrivial task. This is especially true for the continuous time kinetic Ising model where the spins are governed by Glauber dynamics \cite{glauber1963time}. With this dynamics the likelihood contains an exponential function related to the 'non--flipping' times and makes analytical manipulations of the posterior distribution of parameters intractable. However, it is possible to compute the likelihood gradient to find the maximum likelihood estimate (MLE) \cite{zeng2013maximum}.

In this paper we will show how the likelihood for the continuous time problem can be 
remarkably simplified by introducing a combination of two sets of auxiliary random variables. The first set of variables are Poisson random variables which 'linearise' the aforementioned exponential term that appears naturally in likelihoods of Poisson point--process models \cite{wilkinson2006stochastic}. These latent variables are related to previous work, where similar variables have been introduced for sampling the intensity function of an inhomogeneous Poisson process \cite{adams2009tractable}. The second set of variables are the so--called P\'olya--Gamma variables, which were introduced into statistics to enable efficient Bayesian inference for logistic regression \cite{polson2013bayesian} and which may not be familiar in the physics community. These variables have also been used recently for Monte Carlo based Bayesian inference of discrete-time Markov models\cite{linderman2015dependent}, model based statistical testing of spike synchrony \cite{scott2015false}, and an expectation--maximization (EM) scheme for logistic regression \cite{scott2013expectation}. 

With these latent variables the model parameters enter the resulting joint likelihood similarly to simple Gaussian models.  We will use this formulation to construct iterative algorithms for a penalised maximum likelihood and for variational Bayes estimators which have simple analytically computable updates. We test our algorithms on artificial data. As an illustrative application we use the Bayes algorithm on data from a simulated recurrent network with conductance--based spiking neurons and show how the model reproduces the statistics of the data and how the obtained Ising parameters reflect the underlying synaptic structure.

The paper is organized as follows: In Sec.~\ref{sec:model} the continuous time kinetic Ising model is introduced  followed by a derivation of its likelihood in Sec.~\ref{sec:likelihood}. In Sec.~\ref{sec: augmentlikelihood} we introduce auxiliary latent variables to simplify the likelihood. In Sec.~\ref{sec:inference} we develop an EM algorithm for maximum likelihood inference and extend it to L1--regularized likelihood maximisation and a variational Bayes approximation. Finally, in Sec.~\ref{sec:results} we apply our method to simulated data generated from an Ising network and from a network of spiking neurons.

\section{\label{sec:model} The Model}
Following Ref.~\cite{zeng2013maximum} in this section, we consider a system of $N$ Ising spins $s_i(t)\in\{-1,1\}$ for $i=1,\ldots,N$. 
We denote the vector of all spins by $\boldsymbol{s}(t)=(s_1(t),\ldots,s_N(t))^\top$.
A spin $i$ is interacting with spin $j$ through a coupling $J_{ij}$. 
We are not assuming symmetry of these couplings: in general, we have $J_{ij} \neq  J_{ji}$.
We will also allow for {\it self couplings} $J_{ii}$. The total field acting on spin $i$ is given by
\begin{equation}
H_i(t)=\theta_i + \sum_{j=1}^NJ_{ij}s_j(t),
\end{equation}
where $\theta_i$ 
denotes the {\it external field}. The Glauber dynamics of the spins is defined 
by asynchronous updates \cite{zeng2013maximum} where in a small time interval $\Delta t$,  spins $i$ are  selected independently with probability $\gamma \Delta t$ for an update; $\gamma>0$ is the update rate. The updated spins are flipped, i.e.,  $s_i(t+\Delta t) = -s_i(t)$ with probability
\begin{equation}\label{eq:kineticIsing}
P_i^{\text{\small{flip}}}(t) = \frac{\exp(-s_i(t) H_i(t))}{2\cosh (H_i(t))}.
\end{equation}
The probability that spin $i$ is not flipped at time $t$ in the interval $\Delta t$ is given by 
$
1-\gamma \Delta t +  \gamma\Delta t\left(1- P_i^{\text{\small{flip}}}(t)\right).
$
Hence, the total probability of a (time--discretised) temporal  sequence $\{\boldsymbol{s}\}_{0:T}$ 
 of spins during a time interval $[0:T]$ is given by
\begin{equation}\label{eq:likelidisc}
\begin{split}
& P(\{\boldsymbol{s}\}_{0:T}\vert \boldsymbol{J}) = \prod_{(i,t)\in F} \left\{\gamma \Delta t \frac{\exp(-s_i(t) H_i(t))}{2\cosh (H_i(t))} \right\}\\
& \times \prod_{(i,t)\in NF}
\left\{1-\gamma \Delta t + \gamma\Delta t \frac{\exp(s_i(t)H_i(t))}{2\cosh(H_i(t))} \right\}.
\end{split}
\end{equation}
Here $F$ denotes the set of pairs $(i,t)$ where spin $i$ was flipped at time $t$.
and $NF$ is the corresponding, complementary set of times and spins where no flips happened.
$\boldsymbol{J}$ stands for the parameters of the model: 
$\boldsymbol{J}\equiv J_{ij}$ for $i,j =1,\ldots N$ and $\theta_i$ for  $i=1,\ldots,N$.

\section{\label{sec:likelihood} Likelihood and Inference}
Our goal is to infer the couplings and external fields 
from observations of complete spin trajectories over a time interval $[0,T]$. We will consider only {\it likelihood} based approaches in this paper. Hence, we need to compute the probability of spin trajectories (\ref{eq:likelidisc}) as a function of parameters, i.e., the so--called likelihood function in continuous time. Taking the limit $\Delta t\to 0 $ in (\ref{eq:likelidisc}) and discarding prefactors which contain $\Delta t$ but are irrelevant for inference (being independent of $\boldsymbol{J}$), the complete--data likelihood function \cite{wilkinson2006stochastic} is found to be 
\begin{equation}\label{eq:completedatalikelihood}
\begin{split}
& {\cal{L}}(\lbrace\boldsymbol{s}\rbrace_{0:T} \vert \boldsymbol{J}) = \prod_{(i,t)\in F} \frac{\exp(-s_i(t)H_i(t))}{2\cosh (H_i(t))} \\
&
\times \prod_{i=1}^N  \exp\left(\gamma \int_0^T  \left\{\frac{\exp(s_i(t)H_i(t))}{2\cosh (H_i(t))} -1\right\}dt\right).
\end{split}
\end{equation}
A maximum likelihood estimate of the parameters $\boldsymbol{J}$ can be obtained 
by a (possibly penalised) gradient ascent approach of this function \cite{zeng2013maximum}. However, a Bayesian inference approach does not seem to be feasible from the expression (\ref{eq:completedatalikelihood}).  For a Bayesian approach one would introduce a prior density $p(\boldsymbol{J})$ of parameters and would infer statistical properties of $\boldsymbol{J}$ using the posterior
density given by
\begin{equation}
p(\boldsymbol{J} \vert \lbrace\boldsymbol{s}\rbrace_{0:T}) = 
\frac{{\cal{L}}(\lbrace\boldsymbol{s}\rbrace_{0:T} \vert \boldsymbol{J}) p(\boldsymbol{J})}
{\int {\cal{L}}(\lbrace\boldsymbol{s}\rbrace_{0:T} \vert \boldsymbol{J}) p(\boldsymbol{J})\; d\boldsymbol{J}},
\end{equation}
from which posterior expectations of parameters would have to be calculated by high--dimensional integrals. Due to the complex dependency of the likelihood on the parameters, the application of well--known techniques such as Monte Carlo sampling, e.g. using a Gibbs sampler, or approximate inference methods such as the variational approach \cite{bishop2006pattern} would not be trivial. We will show in the next section that the dependency of the likelihood on $\boldsymbol{J}$ can be remarkably simplified by augmenting the system by two sets of auxiliary random variables.

\section{\label{sec: augmentlikelihood} Variable augmentation and tractable likelihood}
The two main problems that prevent us from performing
efficient analytical inference using Eq.~\eqref{eq:completedatalikelihood} come from two sources:
first, the time integral which contains the parameters $\boldsymbol{J}$, 
appears in an exponential function, and, second, the parameters also
appear in the denominators in the hyperbolic cosine function. We will show that both problems can be solved by
the introduction  of auxiliary variables. We will start with a simplification of the integral.
\subsection{Poisson variables}
We note, that fields $H_i(t)$ are piecewise constant functions of time and do not 
change where no spin is flipped. Hence, the time integral can be calculated analytically. We will order the constant intervals and number them by $n\in \{0,1,\ldots, n_{max}\}$. We define $H_i^n$ and $s_i^n$ as the values of the field and spin $i$ between time points $t_n$ and $t_{n+1}$. $t_{n}$ denotes the time of the $n^{\rm{th}}$ flip time for $n\in\{1,\ldots,n_{max}\}$, while $t_0=0$ and $t_{n_{max}+1}=T$. Hence, we obtain
\begin{equation}
\int_0^T \frac{\exp(s_i(t)H_i(t))}{2\cosh (H_i(t))}dt = \sum_{n=0}^{n_{max}}
\frac{\exp(s_i^n H_i^n)}{2\cosh(H_i^n)} ({t_{n+1}} - {t_{n}}).
\end{equation}
Introducing a set of independent Poisson distributed random
variables $\rho_i^n$ for each $i$ and each time slice between $t_{n+1}$ and $t_n$, 
we obtain the following representation of the second part of the likelihood:
\begin{equation}\label{eq:introrho}
\begin{split}
& \exp\left(\gamma \int_0^T  \left\{\frac{\exp(s_i(t)H_i(t))}{2\cosh (H_i(t))} -1\right\}dt\right) \\
& = \prod_{n=0}^{n_{max}}  \left\{\sum_{\rho_i^n = 0}^\infty 
\left(\frac{\exp(s_i^n H_i^n)}{2\cosh (H_i^n)} \right)^{\rho_i^n} 
\Po\left(\rho_i^n | \gamma({t_{n+1}} - {t_{n}})\right)\right\},
\end{split}
\end{equation}
where
\begin{equation}
\Po(\rho \vert \zeta) = 
e^{-\zeta}
\frac{\zeta^{\rho}   }{\rho !},
\end{equation}
denotes a Poisson distribution with mean parameter $\zeta$. For Eq.~\eqref{eq:introrho} we made use of the equality
\begin{equation*}
e^{\zeta(x - 1)} = \sum_{\rho=0}^\infty x^\rho \Po(\rho\vert \zeta),
\end{equation*}
which is the moment--generating function of the Poisson distribution \cite{kingman1992poisson}. Similar variables were used in Ref.~\cite{adams2009tractable} to make Poisson--process likelihoods tractable for Monte--Carlo sampling.
\subsection{P\'olya-Gamma variables}
To get rid of the hyperbolic terms in the denominators, we will use a remarkable representation
which was discovered and used in the statistics literature in recent years to simplify Bayesian inference for logistic regression. Reference \cite{polson2013bayesian} found a convenient form of writing an inverse hyperbolic cosine as a continuous mixture of Gaussian densities as
\begin{equation}
\cosh^{-b}(x) = \int_0^\infty d\omega\ e^{-2\omega x^2}\PG(\omega\vert b, 0),
\end{equation}
where $\PG(\omega\vert b, 0)$ is the {\it P\'olya-Gamma density} with parameter $b$. Surprisingly, the exact form of this distribution is not of importance for our inference algorithm, but only the fact that one can derive its first moments straightforwardly (see Appendix~\ref{app:polya}). Introducing P\'olya-gamma variables $\omega$ into the likelihood~\eqref{eq:introrho} yields the representation
\begin{equation}
p(\lbrace\boldsymbol{s}\rbrace_{0:T}\vert \boldsymbol{J}) = \sum_{\boldsymbol{\rho}} \int 
 {\cal{L}}(\lbrace\boldsymbol{s},\boldsymbol{\rho},\boldsymbol{\omega}\rbrace_{0:T} \vert \boldsymbol{J}) 
 d\boldsymbol{\omega},
\end{equation}
with the augmented likelihood
\begin{equation}\label{eq:augmentedlikelihood}
\begin{split}
& {\cal{L}}(\lbrace\boldsymbol{s}, \boldsymbol{\rho},\boldsymbol{\omega}\rbrace_{0:T} \vert \boldsymbol{J}) = \\
&\prod_{(i,t)\in F}\exp\left[ - s_i(t)H_i(t) - 2 (H_i(t))^2\omega_i(t) \right] \PG(\omega_i(t)\vert 1, 0) 
\\
& \times 
\prod_{i,n} \left(\exp\left[ \rho_i^n (s_i^n H_i^n - \ln(2)) - 2 (H_i^n)^2\omega_i^n \right]\right. \\
& \left. \times \Po\left(\rho_i^n | \gamma({t_{n+1}} - {t_{n}})\right) \PG(\omega_i^n\vert \rho_i^n, 0)\right).
\end{split}
\end{equation}
The advantage of the augmented likelihood over the original one is the fact that
the parameters appear at most quadratically in the exponential functions [note, that the fields $H_i(t)$ are linear functions of the parameters]. As we will see, the computation of maximum likelihood and related estimators as well as Bayesian inference become considerably facilitated.
We will postpone explicit results of Gibbs sampling algorithms to a future publication and discuss applications of the augmented likelihood to penalised maximum likelihood estimation and to 
a variational Bayes algorithm in this paper.

\section{\label{sec:inference} Inference}
\subsection{EM algorithm}
The EM algorithm \cite{dempster1977maximum} is a convenient way to maximise the likelihood iteratively with respect to $\boldsymbol{J}$ by using latent variable representations. The algorithm cycles between an E--step and an M--step and guarantees to increase the likelihood~\eqref{eq:completedatalikelihood} in each step. At iteration $m+1$, in the 

\noindent
{\bf E-step} one computes the cost function $Q(\boldsymbol{J}, \boldsymbol{J}_m)$. It  equals 
the expectation of the logarithm of the augmented likelihood with respect to the
distribution of latent variables conditioned on the parameters at the previous iteration $m$
\begin{equation}
\begin{split}
& Q(\boldsymbol{J}, \boldsymbol{J}_m) \doteq \\
& \sum_{\boldsymbol{\rho}} \int d\boldsymbol{\omega}\; p(\boldsymbol{\rho},\boldsymbol{\omega} \vert \{\boldsymbol{s}\}_{0:T}, \boldsymbol{J}_m)
\ln {\cal{L}}(\lbrace\boldsymbol{s},\boldsymbol{\rho},\boldsymbol{\omega}\rbrace_{0:T}\vert \boldsymbol{J}).
\end{split}
\end{equation}
\noindent
{\bf M--step} Here we compute an update of the parameters via
\begin{equation}
\boldsymbol{J}_{m+1} = \arg\max_{\boldsymbol{J}} Q(\boldsymbol{J}, \boldsymbol{J}_m).
\end{equation}
The conditional distribution is given by
\begin{equation}
\begin{split}
& p(\{\boldsymbol{\rho},\boldsymbol{\omega}\}_{0:T} \vert \{\boldsymbol{s}\}_{0:T}, \boldsymbol{J}) = \\
&
p(\{\boldsymbol{\omega}\}_{0:T} \vert \{\boldsymbol{s}, \boldsymbol{\rho}\}_{0:T},\boldsymbol{J})
P(\{\boldsymbol{\rho}\}_{0:T} \vert \boldsymbol{J},\{\boldsymbol{s}\}_{0:T}),
\end{split}
\end{equation}
where
\begin{equation}
\begin{split}
p(\{\boldsymbol{\omega}\}_{0:T} \vert \{\boldsymbol{s}, \boldsymbol{\rho}\}_{0:T},\boldsymbol{J}) = &
 \prod_{(i,t)\in F} \PG(\omega_i(t)\vert 1, 2H_i(t)) \\
& \times \prod_{n,i}\PG(\omega_i^n\vert \rho_i^n, 2H_i^n),
 \end{split}
\end{equation} 
where we defined the {\em tilted P\'olya--Gamma} distribution as 
$$
\PG(\omega_i^n\vert b, c) = \frac{\exp\left( - \frac{c^2}{2}\omega_i^n \right)\PG(\omega_i^n\vert b, 0)}{\cosh^{-b}(c/2)},
$$
and where
\begin{equation}
P(\boldsymbol{\rho} \vert \boldsymbol{J},\{\boldsymbol{s}\}_{0:T})=  \prod_{n,i}
\Po\left(\rho_i^n\left\vert \gamma(t_{n+1} - t_n)\frac{\exp(s_i^n H_i^n)}{2\cosh (H_i^n)}\right.\right).
\end{equation}
The first part of the conditional density is over factorising P\'olya Gamma variables and the second one over factorising Poisson random variables. The necessary expectations for the E--step follow from simple properties of Poisson random variables
and of P\'olya Gamma random variables derived in Appendix~\ref{app:polya}. This results in
\begin{equation}
\begin{split}
& \langle\omega_i(t) \rangle = \frac{1}{4 H_i(t)}\tanh(H_i(t)), \\
& \langle\omega_i^{n} \rangle = \frac{\langle\rho_i^{n}\rangle}{4 H_i^{n}}\tanh(H_i^{n}),\\
& \langle \rho_i^n\rangle = (t_{n+1} - t_n) \gamma \frac{\exp(s_i^n H_i^n)}{2\cosh(H_i^n)},
\end{split}
\end{equation}
where the brackets $\langle\cdot\rangle$ denote expectations conditioned on $\boldsymbol{J}_m$.
Since the augmented log--likelihood is a quadratic form in the parameters $\boldsymbol{J}$, the 
maximisation leads to $N$ systems of linear equations 
for the vectors $\boldsymbol{J}_{i\cdot} \doteq (\theta_i, J_{i1}, \ldots, J_{iN})^\top$
of the form
\begin{equation}\label{eq:linearsystem}
A_i\boldsymbol{J}_{i\cdot} = \boldsymbol{b}_{i\cdot},
\end{equation}
with
\begin{equation}
b_{ij} = -\sum_{t\in F(i)} s_i(t) s_j(t) + \sum_{n} \langle \rho_i^n\rangle s_i^n s_j^n,
\end{equation}
and 
\begin{equation}
A_{ijk} = 4\left(\sum_{t \in F(i)} \langle\omega_i^{t}\rangle s_k(t) s_j(t)+ \sum_{n} \langle\omega_i^n \rangle s_k^n s_j^n \right).
\end{equation}
Here $F(i)$ is the set of all times that spin $i$ has flipped. As mentioned before only the first moment of the P\'olya--Gamma density is required.

\subsection{\label{subsec:sparsity} Sparsity via L1 regularization}
Assuming a factorising Laplace distribution over each coupling $J_{ij}$ 
$$
p(J_{ij}) = \frac{\lambda}{2}\exp\left(-\lambda|J_{ij}|\right),
$$
will enforce sparsity on the network. $\lambda$ is the scale parameter of this density.
On the level of the MAP (maximum{\it a posteriori}) Bayesian estimator this is equivalent to L1 regularised maximum likelihood estimation. However, the absolute value in the exponent of this prior would prevent us from using the previously described EM procedure
directly and allow only for gradient methods similar to Ref.~\cite{zeng2014l1}. Fortunately, this problem can again be solved by the introduction of a further auxiliary random variable
for each single coupling parameter $J_{ij}$. This follows from the fact that a Laplace distribution can once more be represented as an infinite mixture of Gaussians \cite{girosi1991models,pontil2000noise},
\begin{equation}
\begin{split}
& \frac{\lambda}{2} \exp(-\lambda | J |) = \\
& \int d\beta \sqrt{\frac{\beta \lambda^2}{2\pi}}\exp\left(-\frac{\beta\lambda^2}{2}J^2\right)p(\beta),
\end{split}
\end{equation}
with 
$$
p(\beta)=(\beta/2)^{-2}\exp\left(-1/(2\beta)\right).
$$
By extending the augmented likelihood~\eqref{eq:augmentedlikelihood} to  {\it sparsity variables} $\{\beta_{ij}\}$ a similar EM algorithm is possible to obtain the L1--regularized ML solution of $\boldsymbol{J}$. The required
conditional density factorises as
\begin{equation}
p(\{\boldsymbol{\rho},\boldsymbol{\omega}\}_{0:T}, \boldsymbol{\beta} \vert \{\boldsymbol{s}\}_{0:T}, \boldsymbol{J}) =
p(\{\boldsymbol{\rho},\boldsymbol{\omega}\}_{0:T}\vert \{\boldsymbol{s}\}_{0:T}, \boldsymbol{J}) \;
p(\boldsymbol{\beta} \vert \boldsymbol{J}),
\end{equation}
where $p(\boldsymbol{\beta} \vert \boldsymbol{J}) = \prod_{i,j} p(\beta_{ij} \vert J_{ij})$  and each factor is a {\it generalized inverse Gaussian} distribution
\begin{equation}
\begin{split}
p(\beta_{ij} \vert J_{ij}) = & \GIG(\beta_{ij}\vert a_{ij}, 1,\nu)\\
= & \frac{a_{ij}^{\nu/2}}{2K_\nu(\sqrt{a_{ij}})}\beta_{ij}^{\nu-1}\exp\left(-\frac{a_{ij} \beta_{ij} +1/\beta_{ij}}{2}\right),
\end{split}
\end{equation}
where $a_{ij} =\lambda^2J_{ij}^2$, $\nu=-1/2$, and $K_\nu$ is the modified Bessel function of the second kind.
The only change in the linear system \eqref{eq:linearsystem} is in the matrices $A$, which have to be replaced by
\begin{equation}
A^{sparse}_{ijk} = A_{ijk}  + \delta_{i,k}\lambda^2\langle \beta_{ij}\rangle,
\end{equation}
and where $\langle \beta_{ij}\rangle = (J_{ij}^2\lambda^2)^{-1/2}$ (see Appendix~\ref{app:sparsity}).

\subsection{\label{subsec:bayesian} Approximate posterior distribution via variational Bayes}
For Bayesian inference we assume the previously discussed Laplace prior over couplings $J_{ij}$ with scaling parameter $\lambda$ and for the external fields $\theta_i$ a Gaussian prior with mean $\mu_{\theta}$ and precision $\lambda_\theta^2$. To obtain a full posterior distribution including the couplings $\boldsymbol{J}$ we could either sample from the posterior or resort to a variational approach. The latter method is popular in the field of machine learning \cite{bishop2006pattern} but has its roots in statistical physics \cite{feynman1998statistical}.
In our case we assume approximated posterior that has the following factorising form:
\begin{equation}\label{eq:variationalposterior}
\begin{split}
& p(\boldsymbol{J},\{\boldsymbol{\omega},\boldsymbol{\rho}\}_{0:T},\boldsymbol{\beta}  \vert \lbrace\boldsymbol{s}\rbrace_{0:T}) \approx   \\
& q(\boldsymbol{J}, \{\boldsymbol{\omega}, \boldsymbol{\rho}\}_{0:T},\boldsymbol{\beta}) \equiv
q_1(\boldsymbol{J}) q_2(\{\boldsymbol{\omega}, \boldsymbol{\rho}\}_{0:T},\boldsymbol{\beta}), 
\end{split}
\end{equation}
where the two factors $q_1$ and $q_2$ are optimised to minimise the relative entropy (Kullback--Leibler)
divergence:
\begin{equation}\label{eq:kl}
\begin{split}
D(q  ; p) = & \sum_{\boldsymbol{\rho}} \left[ \int q(\boldsymbol{J}, \{\boldsymbol{\omega}, \boldsymbol{\rho}\}_{0:T},\boldsymbol{\beta}) \right. \\
& \left. \times \ln \frac{q(\boldsymbol{J}, \{\boldsymbol{\omega}, \boldsymbol{\rho}\}_{0:T},\boldsymbol{\beta})}{p(\boldsymbol{J},\{\boldsymbol{\omega},\boldsymbol{\rho}\}_{0:T},\boldsymbol{\beta}  \vert \lbrace\boldsymbol{s}\rbrace_{0:T}) } d\boldsymbol{\omega} d\boldsymbol{\beta}d\boldsymbol{J}\right].
\end{split}
\end{equation}
This is equivalent to minimising the {\it variational free energy}
\begin{equation}\label{eq:free energy}
\begin{split}
\mathcal{F}(q;p) = & \sum_{\boldsymbol{\rho}}\left[ \int q(\boldsymbol{J}, \{\boldsymbol{\omega}, \boldsymbol{\rho}\}_{0:T},\boldsymbol{\beta})\right. \\
& \left. \times \ln \frac{q(\boldsymbol{J}, \{\boldsymbol{\omega}, \boldsymbol{\rho}\}_{0:T},\boldsymbol{\beta})}{p(\lbrace\boldsymbol{s},\boldsymbol{\omega},\boldsymbol{\rho}\rbrace_{0:T}, \boldsymbol{J},\boldsymbol{\beta}) }d\boldsymbol{\omega} d\boldsymbol{\beta}d\boldsymbol{J}\right].
\end{split}
\end{equation}
The negative free energy is actually a lower bound on the log marginal likelihood
\begin{equation}
-\mathcal{F}(q;p) \leq \int \mathcal{L}(\{\boldsymbol{s}\}_{0:T}\vert \boldsymbol{J})p(\boldsymbol{J})d\boldsymbol{J},
\end{equation}
and can be used directly for approximate model selection \cite{bishop2006pattern}, while in a pure maximum likelihood approach this is not possible. 

Minimising the variational free energy with respect to the factors of our factorizing distribution~\eqref{eq:variationalposterior}, the optimal factors turn out to be
\begin{align*}
& q_1^\star(\boldsymbol{J}) \propto \exp\left(\left\langle \ln p(\boldsymbol{J},\{\boldsymbol{s},\boldsymbol{\omega},\boldsymbol{\rho}\}_{0:T},\boldsymbol{\beta}) \right\rangle_{q_2}\right), \\
& q_2^\star(\{\boldsymbol{\omega},\boldsymbol{\rho}\}_{0:T},\boldsymbol{\beta}) \propto \exp\left(\left\langle \ln p(\boldsymbol{J},\{\boldsymbol{s},\boldsymbol{\omega},\boldsymbol{\rho}\}_{0:T},\boldsymbol{\beta}) \right\rangle_{q_1}\right),
\end{align*}
which are obtained by iterative updates \cite{bishop2006pattern}. For the posterior at hand we find the optimal factor $q_2$ of the posterior
\begin{equation}
\begin{split}
& q_2^\star(\boldsymbol{\rho},\boldsymbol{\omega}, \boldsymbol{\beta}) = \prod_{(i,t)\in F}q_2(\omega_i(t))\prod_{i,n}q_2(\omega_i^n\vert \rho_i^n)q_2(\rho_i^n)
\; q_2(\boldsymbol{\beta})\\
= & \prod_{(i,t)\in F} \PG(\omega_i(t)\vert 1, 2\sqrt{\langle (H_i(t))^2\rangle}) \\
& \times \prod_{i,n}\PG\left(\omega_i^n\vert \rho_i^n, 2\sqrt{\langle (H_i^n)^2\rangle}\right) \\
& \times \Po\left(\rho_i^n\left\vert \gamma(t_{n+1} - t_n)\frac{\exp(s_i^n \langle H_i^n\rangle)}{2\cosh \left(\sqrt{\langle (H_i^n)^2\rangle}\right)}\right.\right)\\
& \times \prod_{(ij)}\GIG(\beta_{ij}\vert \langle J_{ij}^2 \rangle\lambda^2,1,-1/2).
\end{split}
\end{equation}
From the fact that the augmented likelihood~\eqref{eq:augmentedlikelihood} and the sparsity prior factorise in the components $\boldsymbol{J}_{i\cdot}$ it follows that the optimal posterior $q_1^\star(\boldsymbol{J})$ does so as well. Each of those factors is a Gaussian distribution with covariance and mean given by
\begin{eqnarray}
& \Sigma_i = \left( 4  A_i + \left(\tilde{\Sigma}_i\right)^{-1}\right)^{-1}, \\
& \boldsymbol{\mu}_i = \Sigma_i \left( b_i + \left(\tilde{\Sigma}_i\right)^{-1}\tilde{\boldsymbol{\mu}}_i\right),
\end{eqnarray}
where $\tilde{\Sigma}_i$ is a diagonal matrix with $diag(\tilde{\Sigma}_i^{-1}) = \left(\lambda_\theta^2, \lambda^2\left\langle \beta_{i1}\right\rangle, \ldots, \lambda^2\left\langle \beta_{iN}\right\rangle\right)$. The prior mean is defined as $\tilde{\boldsymbol{\mu}}_i=(\mu_\theta,0,\ldots,0)^\top$. Similar to the EM algorithm, we have a variational step, where $q_2$ is optimised, given $q_1$ and a second one, optimising $q_1$ given $q_2$. The variational step updating $q_2$ differs from E--step in the sense, that here expectations over the terms with the couplings $\boldsymbol{J}$ are required and not only the pointwise estimate (see Appendix~\ref{app:variational}). The variational M--step is similar to the EM algorithm, where the expectations for $A$ and $b$ are computed with respect to $q_2$.
\par
The Python code of the algorithms discussed here is publicly available \footnote{Python code:\newline \href{https://github.com/christiando/dynamic_ising.git}{\nolinkurl{https://github.com/christiando/ dynamic_ising.git}}}. 
\section{\label{sec:results} Results}
\begin{figure}
\includegraphics{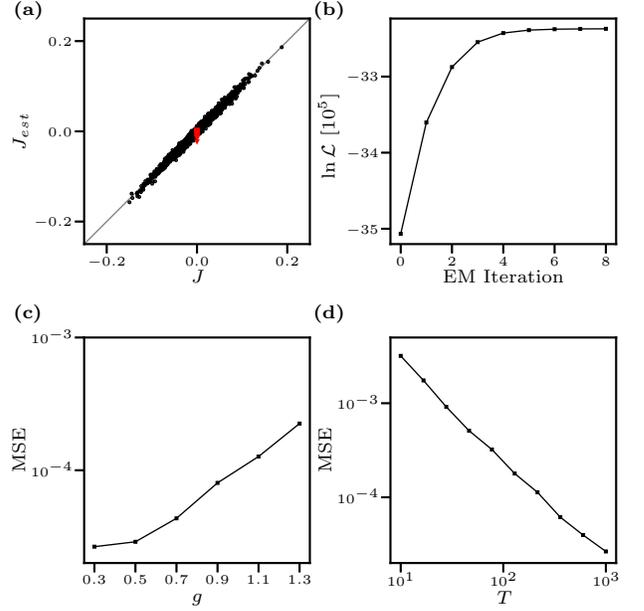}
\caption{Inference with EM algorithm on artificial data. {\bf (a)} True couplings (black dots) and external fields (red triangles) vs. inferred ones. {\bf (b)} The log--likelihood as function of EM iterations. The parameters are set to $N=40$, $T=10^3$, and $g=0.3$ with external fields $\boldsymbol\theta=\boldsymbol{0}$. {\bf (c)} MSE between $\boldsymbol{J}$ and $\boldsymbol{J}_{est}$ as a function of scaling factor of the variance $g$ and {\bf (d)} as a function of data length $T$. If not changed parameters are as in (a).
}\label{fig:fig1}
\end{figure}
We test the EM algorithm on artificial data generated with random couplings $J_{ij}$ from a Gaussian distribution with mean $0$ and variance $g^2/N$, where scaling factor $g=0.3$. With external fields $\boldsymbol\theta=\boldsymbol{0}$ and update rate $\gamma=100$ data is generated with a Gillespie--algorithm \cite{wilkinson2006stochastic} (see Appendix~\ref{app:sampling}).
\subsection{Maximum likelihood}
In Fig.~\ref{fig:fig1} the inference results for the EM algorithm are shown. Figures~\ref{fig:fig1}(a) and~\ref{fig:fig1}(b) present a single fit with $N=40$ spins and data length $T=10^3$. The inferred couplings $\boldsymbol{J}_{est}$ agree well with the true couplings $\boldsymbol{J}$. The logarithm of the likelihood \eqref{eq:completedatalikelihood} converges well after eight EM iterations. The mean squared error (MSE) increases with increasing scaling of the coupling variance $g$ [Fig.~\ref{fig:fig1} {\bf (c)}] and decreases linearly on a log-log scale with increasing data length $T$ [Fig.~\ref{fig:fig1} {\bf (d)}].
\begin{figure}
\includegraphics{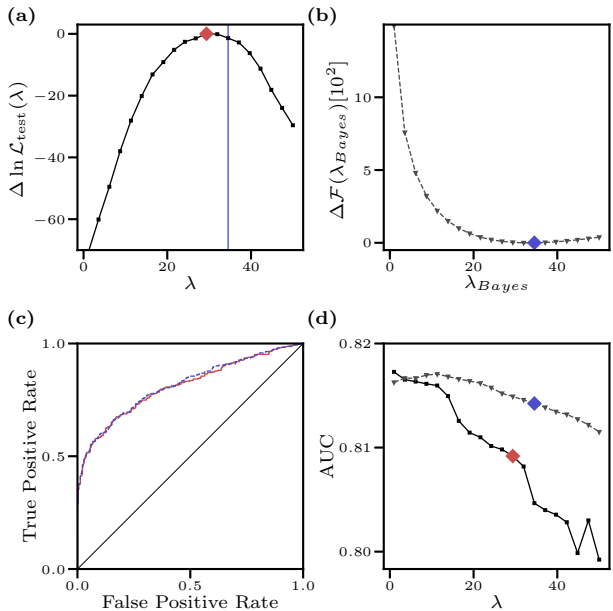}
\caption{Inference of sparse couplings with EM and variational Bayes. Artificial data ($T=50,N=25,g=0.3$) are generated, but each coupling is set to $0$ with probability $p_{\mathrm{sparse}}=1/2$. {\bf (a)} Difference in likelihood (with respect to the likelihood obtained with optimal $\lambda^\star$) of couplings $\boldsymbol{J}_{est}$ inferred by EM as a function of regularization parameter $\lambda$. Likelihood $\mathcal{L}_{\rm{test}}$ is computed on unseen test data ($T=50$). The optimal parameter is $\lambda^\star=29.4$ (red diamond). The vertical line marks the variational estimation $\lambda^\star_{Bayes}$. {\bf (b)} Difference in free energy $\mathcal{F}$ (with respect to the likelihood obtained with estimate of optimal $\lambda^\star_{Bayes}$) of the variational Bayes algorithm. The optimal parameter is $\lambda_{Bayes}^\star=34.5$ (blue diamond). {\bf (c)} ROC curves for the $\lambda^\star$ (EM, solid red line) and $\lambda_{Bayes}^\star$ (Bayes, dashed blue line), respectively. {\bf (d)} The AUC for different parameters $\lambda$ for the EM result (solid black line with squares) and the variational Bayes algorithm (dashed gray line with triangles). Diamonds mark the optimal $\lambda^\star$ and the estimate $\lambda^\star_{Bayes}$.}\label{fig:fig2}
\end{figure}
\subsection{L1 Regularization and Variational Bayes}
Regularization becomes particular important once little data are at hand. To test this we generate couplings as before for a network of $N=25$ spins, but a coupling is set to $0$ with probability of $p_{\mathrm{sparse}}=0.5$. Generated data have length $T=50$. We run the L1--regularized EM algorithm with different values of $\lambda$ and define the optimal $\lambda^\star$, whose MLE $\boldsymbol{J}_{est}$ maximises the likelihood $\mathcal{L}_{\rm{test}}$ on unseen test data ($T=50$) generated by the true Ising parameters $\boldsymbol{J}$ [see Fig.~\ref{fig:fig2} {\bf (a)}]. For inference by the variational algorithm on the same training data we estimate the optimal $\lambda^\star_{Bayes}$ by taking the value that minimises the free energy \eqref{eq:free energy} [Fig.~\ref{fig:fig2} {\bf (b)}]. Note, that the Bayesian algorithm requires no test data for this estimate.
\begin{figure}
\includegraphics{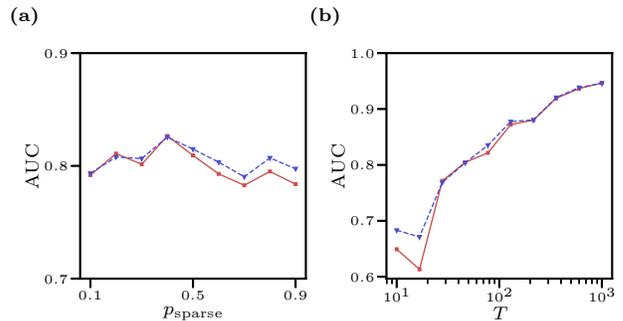}
\caption{The classification of non--zero couplings depending on sparsity of couplings and data length. {\bf (a)} The AUC depending on the sparsity, i.e. the probability of a true coupling being $0$, and in {\bf (b)} depending on length of training data $T$. Results for EM shown by the red solid and variational Bayes algorithm by the blue dashed line. If not changed, parameters are as in Fig. ~\ref{fig:fig2}.}\label{fig:fig_supp}
\end{figure}
\par
Next we try to find the nonzero couplings from our fitting results. For the L1--penalized MLE an estimated coupling is considered as nonzero if $\vert J_{ij}^{est}\vert\geq z$, where the $z$ is an arbitrary threshold. To make use of the additional information of uncertainty, for the variational Bayes couplings are considered to be nonzero if $\vert \langle J_{ij}\rangle_{q_1}\vert \geq z\sqrt{(\Sigma_{i})_{(jj)}}$. The classification of nonzero-couplings is quantified by plotting the false positive rate (proportion of zero couplings that are misclassified as nonzero) versus the true positive rate (proportion of nonzero couplings that are correctly classified as nonzero) for a varying threshold $z\in[0,\infty]$. This is the Receiver--Operator characteristic (ROC) curve (see Fig.~\ref{fig:fig2} {\bf (c)} for $\lambda^\star$ and $\lambda_{Bayes}^\star$ respectively). As a measure of classification performance we use the area under the ROC curve (AUC), which is $1$ for perfect classification and $1/2$ at chance level. Figure~\ref{fig:fig2}{\bf (d)} shows that performance for the EM and the variational Bayes algorithm differ only marginally. For both algorithms the AUC is approximately constant, when repeating the same data generating and fitting procedure as before, but with varying sparsity $p_{\rm{sparse}}$ [Fig.~\ref{fig:fig_supp} {\bf(a)}]. When increasing the length of training data $T$ the AUC increases as expected [Fig.~\ref{fig:fig_supp} {\bf(b)}]. For subsequent analysis we will focus on the variational Bayes algorithm.

\subsection{Inference of biophysical network}
As an application of our algorithm we fit our model to data generated from a more biologically plausible network. We simulate a recurrent network of $1000$ leaky integrate--and--fire neurons ($800$ excitatory and $200$ inhibitory neurons) receiving Poisson input (see Appendix~\ref{app:network} and Ref.~\cite{renart2010asynchronous}). The synapses connect neurons randomly, are conductance--based, and vary in strengths and delays. The network is simulated for $T=1000\rm{s}$. Spike times of $30$ excitatory and $10$ inhibitory neurons are used for fitting the kinetic Ising model, where neuron $i$ is considered as 'active' for $10\rm{ms}(=\gamma^{-1})$ after each spike $s_i(t) = 1$ and 'inactive' otherwise [$s_i(t)=-1$]. The two questions we address here are, (1) how well does the fitted model reproduce the statistics of the recorded data, and (2) how are the synapses reflected in the estimated coupling parameters $\boldsymbol{J}$?

For the first question we compare data obtained from the spiking network with data sampled from the fitted the kinetic Ising model $\langle \boldsymbol{J}\rangle_{q_1}$ $(T=1000 \rm{s})$. To compare the original data with the Ising model data the (second--order) correlations from these data are computed as
\begin{equation}\label{eq:correlation}
C_{ij} = \frac{1}{T}\int_0^T (s_i(t) - m_i)(s_j(t) - m_j)dt,
\end{equation}
where the mean is given by $m_i = \int_0^T s_i(t) dt/T$.
\begin{figure}
\includegraphics{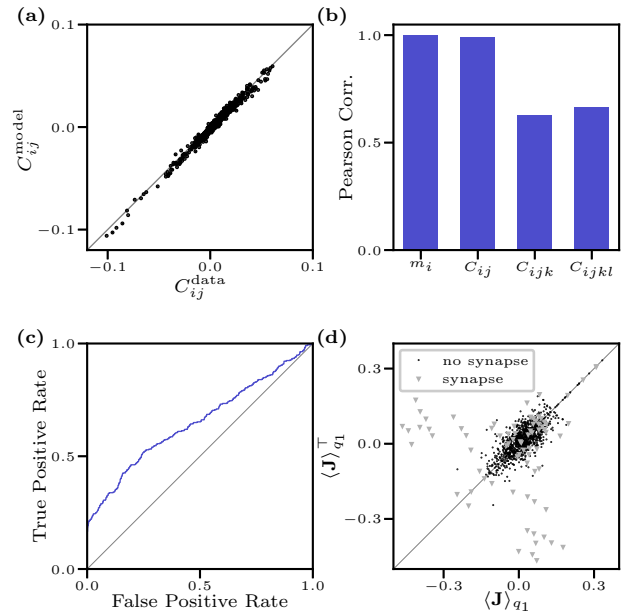}
\caption{Model fitted to data from a simulated recurrent network. {\bf (a)} Second--order correlation $C_{ij}$ of the original data vs. data sampled from mean couplings $\langle \boldsymbol{J}\rangle_{q_1}$ obtained via variational Bayes. {\bf (b)} Pearson correlation between first to fourth order correlations of real sampled data. {\bf (c)} ROC curve for identifying synapses with posterior over couplings $\boldsymbol{J}$ ($\rm{AUC}=0.65,\lambda^\star_{Bayes}=16.5$). {\bf (d)} Mean couplings $\langle \boldsymbol{J}\rangle_{q_1}$ of the variational posterior vs. transpose. Couplings between neurons connected by a synapse are marked with gray triangles.}\label{fig:fig3}
\end{figure}

The results are compared in Fig.~\ref{fig:fig3}{\bf (a)}, and we find good agreement of original data and Ising samples. Furthermore, we compute the higher order correlations $C_{ijk}$ and $C_{ijkl}$ of the data and calculate the Pearson correlation coefficient between correlations from the original and the sampled data [see Fig.~\ref{fig:fig3}{\bf (b)}]. The first two correlations ($m_i$ and $C_{ij}$) yield a Pearson correlation close to $1$. Interestingly the Pearson correlation coefficient is strongly positive for $C_{ijk}$ and $C_{ijkl}$ as well, indicating that the Ising model also carries information about higher order correlations in the data. 

As before we try to identify synapses in the simulated network by ROC curve analysis (Fig.~\ref{fig:fig3} {\bf (c)}). The classification yields an $\rm{AUC}=0.65$ ($\lambda^\star_{Bayes}=16.5$). Even though there is information about the synapses, many more nonzero couplings are estimated that do not directly reflect synapses in the network. This is possibly caused by the fact, that the network is only partially observed and the kinetic Ising model compensates for this part with more nonzero couplings.

Previous work has indicated for the kinetic Ising model in discrete time \cite{hertz2011ising,zeng2013maximum}, that for experimental data recorded {\it in vivo} the estimated couplings $\boldsymbol{J}$ show a symmetric signature: $J_{ij}\approx J_{ji}$. This is particularly interesting for the Ising model in continuous time, since for the model with symmetric couplings the stationary distribution is given by the maximum entropy equilibrium model \cite{glauber1963time} and potentially justifies the use of static Ising models for such data. As an indicator for symmetry we plot the mean of the variational posterior obtained from the recurrent network versus its transpose [Fig.~\ref{fig:fig3} {\bf (d)}]. We observe, that many couplings are indeed close to the diagonal, while some show large deviations from it. However, those with strong deviations correspond to the couplings which reflect synapses in the underlying network. Hence, the approximately symmetric part is not caused by synapses, but either by our data transformation to fit the Ising model or by the fact that we only partially observe the network. 
\section{\label{sec:discussion} Discussion}
In this paper we have presented efficient algorithms for inferring the couplings of a continuous time kinetic Ising model defined by Glauber dynamics. Using a combination of two auxiliary latent variable sets the complete data log--likelihood becomes a simple quadratic function in the couplings. A third set of auxiliary variables allows us to deal with sparse couplings, equivalent to an L1--penalized likelihood without resorting to gradient--based algorithms \cite{zeng2014l1}. Using this representation we derive an EM algorithm for (penalised) maximum likelihood estimation of the couplings with explicit analytical updates. This leads to a guaranteed increase of the likelihood in each iteration. The computational complexity is similar to a Newton--Raphson method for optimising the original log--likelihood, since the Hessian matrix requires a similar inverse of the summed data covariances \cite{zeng2013maximum}. However, our algorithm does not require any tuning of a step size.

We have extended our latent variable approach to a Bayesian scenario but 
have restricted ourselves to a fast variational Bayes approximation.
However, it is straightforward to develop a Monte Carlo Gibbs sampler for the latent variable structure. This would require drawing samples from P\'olya--Gamma density rather than computing only its mean. We have tested our inference algorithms on simulated data demonstrating fast convergence of the method. The variational Bayes approximation allows us to perform model selection, yielding hyper--parameters which achieve close to optimal likelihoods on test data. As an application of our approach we have investigated the quality of the kinetic Ising model to describe data which were generated from a more realistic, biologically inspired {\em integrate and fire} neural network model which is only partially observed. We have shown that the kinetic Ising model reproduces low order statistics of the data well. However, the partial observation of neurons prohibits a safe identification of synapses in terms of the Ising coupling parameters. It would be interesting to see if the performance of a kinetic Ising model on such data could be improved by including explicit unobserved neurons and their couplings in the model \cite{roudi2015learning}.  We expect that our latent variable approach would facilitate statistical inference for such an extended model and provide alternatives to current approximate inference methods \cite{tyrcha2013network,dunn2013learning,bachschmid2014inferring,battistin2015belief}. We are currently working on an extension of our inference approach by including time--dependent model parameters which makes the model more realistic and which has been shown of importance for biological data analysis \cite{tyrcha2013effect,donner2017approximate}.

Finally, our latent variable approach should also be applicable to other inference problems for point process models; e.g., a combination with Gaussian process priors should allow for nonparametric approximate inference of rate functions for inhomogeneous Poisson processes \cite{adams2009tractable}. Models with similar point--process likelihoods are common in neuroscience \cite{brillinger1988maximum,paninski2004maximum,latimer2014inferring}, for modeling seismic activity \cite{ogata1998space}, analyzing social network analysis \cite{zhao2015seismic}, etc.

\section{Acknowledgements}
C.D. was supported by the Deutsche Forschungsgemeinschaft (GRK1589/2).

\appendix
\section{\label{app:sampling}Generating Data}
To generate artificial data for the kinetic Ising model in continuous time we can make use of the Gillespie algorithm \cite{wilkinson2006stochastic}. Having a coupling matrix $\boldsymbol{J}$ for $N$ spins and an initial data vector $\boldsymbol{s}(0)$ data are generated as follows: (1) We draw the next update time $t^\prime$ from a exponential distribution with mean $(\gamma\times N)^{-1}$, (2) we draw a spin $i$ with probability $1/N$, and finally (3) we flip spin $i$ at time $t^\prime$ according to Eq.~\eqref{eq:kineticIsing} and set $t\leftarrow t^\prime$. These three steps are repeated until $t\geq T$.

\section{\label{app:polya}Properties of P\'olya--Gamma distribution}
The  P\'olya--Gamma density \cite{polson2013bayesian} allows us to represent the inverse hyperbolic cosine function as an infinite Gaussian mixture
\begin{equation}\label{eq:pg cosh}
\cosh^{-b}(c/2) = \int_0^\infty d\omega \exp(-\frac{c^2}{2}\omega)\PG(\omega\vert b, 0).
\end{equation}
Furthermore, we define the {\it tilted P\'olya--Gamma distribution} as
\begin{equation}\label{eq:tilted pg}
\PG(\omega\vert b, c) \propto e^{-c^2/2\omega}\PG(\omega\vert b, 0).
\end{equation}
From Eqs.~\eqref{eq:pg cosh} and~\eqref{eq:tilted pg} we obtain the moment generating function
\begin{equation}\label{eq:polyagammalaplace}
\begin{split}
\langle e^{\omega t}\rangle = & \frac{\cosh^b(c/2)}{\cosh^b\left(\sqrt{\frac{c^2/2 - t}{2}}\right)}.
\end{split}
\end{equation}
By differentiating \eqref{eq:polyagammalaplace} at $t=0$ the analytical form of the expectation of $\omega$ is obtained
\begin{equation}
\langle \omega \rangle = \frac{b}{2c}\tanh\left(\frac{c}{2}\right).
\end{equation}

\section{\label{app:sparsity}Latent variable representation of Laplace distribution}
The Laplace distribution can written as an infinite mixture of Gaussians \cite{girosi1991models,pontil2000noise}
\begin{equation}
 \frac{\lambda}{2}\exp(-\lambda\vert x\vert) = \int_0^\infty \sqrt{\frac{\beta\lambda^2}{2\pi}}\exp\left(-\frac{\beta\lambda^2}{2} x^2\right)p(\beta)d\beta,
\end{equation}
with
\begin{equation}
p(\beta) = (\beta/2)^{-2}\exp\left(-\frac{1}{2\beta}\right).
\end{equation}
By inspection we find the conditional density
\begin{equation}
p(\beta\vert x) = \GIG(\beta\vert x^2\lambda^2, 1, -1/2),
\end{equation}
where $\GIG$ is a generalized inverse Gaussian distribution defined as
\begin{equation}
\GIG(\beta\vert a, b, \nu) = \frac{(a/b)^{\nu/2}}{2K_\nu(\sqrt{ab})}\beta^{\nu-1}\exp\left(-(a\beta - b/\beta)/2\right),
\end{equation}
and $K_\nu$ is the modified Bessel function of the second kind. The expectations of $\beta$ are 
\begin{equation}
\langle \beta \rangle = \frac{K_{1/2}(\sqrt{x^2\lambda^2})}{\sqrt{x^2\lambda^2}K_{-1/2}(\sqrt{x^2\lambda^2})} = \frac{1}{\sqrt{x^2\lambda^2}},
\end{equation}
where the Bessel functions cancel due to $K_{\nu}(\sqrt{x^2\lambda^2})=K_{-\nu}(\sqrt{x^2\lambda^2})$.

\section{\label{app:variational}Variational Bayes}
In the variational Bayes algorithm the updates in the step updating $q_2$ involve the expectations $\langle H_i^{t,n}\rangle_{q_1}$ and $\langle (H_i^{t,n})^2\rangle_{q_1}$ instead of only the pointwise MLE in the E--step of the EM algorithm. The required expectations are
\begin{equation}
\begin{split}
& \langle\omega_i(t) \rangle = \frac{1}{4 \sqrt{\langle (H_i(t))^2 \rangle}}\tanh(\sqrt{\langle (H_i(t))^2 \rangle}),\\
& \langle\omega_i^{n} \rangle = \frac{\langle\rho_i^{n}\rangle}{4 \sqrt{\langle (H_i^n)^2 \rangle}}\tanh(\sqrt{\langle (H_i^n)^2 \rangle}),\\
& \langle \rho_i^n\rangle = (t_{n+1} - t_n) \gamma \frac{\exp(s_i^n \langle H_i^n\rangle)}{2\cosh(\sqrt{\langle (H_i^n)^2 \rangle})}.
\end{split}
\end{equation}
The free energy \eqref{eq:free energy}, that is minimised in the variational Bayes algorithm is easy to calculate since we immediately see that the terms involving $\PG(\omega_i(t)\vert 1, 0)$, $\PG(\omega_i^n)\vert \rho_i^n, 0)$, $\Po(\rho_i^n\vert \gamma(t_{n+1}-t_n))$ and $p(\beta)$ appear in the nominator as well as in the denominator and cancel out. The free energy at a minimum is
\begin{equation}
\begin{split}
\mathcal{F}(q^\star;p) = & \sum_{(i,t)\in F} \ln\frac{2\cosh\left(\sqrt{\langle(H_i(t))^2\rangle}\right)}{\exp(-s_i(t)\left\langle H_i(t)\right\rangle)} \\
& + \sum_{i,n}\gamma(t_{n+1}-t_n)\left(1-\frac{\exp\left(s_i^n\left\langle H_i^n\right\rangle\right)}{2\cosh(\sqrt{\langle(H_i^n)^2\rangle})}\right) \\
& + \sum_{i,j}\ln \left(\frac{\sqrt{2\pi}\langle J_{ij}^2\rangle^{-1/4}}{(2\sqrt{\lambda})^3K_{-1/2}(\sqrt{\lambda^2\langle J_{ij}^2\rangle})}\right) \\
& - \sum_i\left\langle \ln \mathcal{N}(\theta_i\vert \mu_\theta, \lambda^{-2}_\theta) \right\rangle  + \left\langle \ln q_1(\boldsymbol{J})\right\rangle,
\end{split}
\end{equation}
where all expectations are taken over the variational posterior $q^\star$. Note the similarity of the first two summands and the likelihood~\eqref{eq:completedatalikelihood}.

\section{\label{app:network} Simulated network of spiking neurons}
We simulate a spiking network similar to the one described in Ref.~\cite{renart2010asynchronous}, Figure 3. The network consisted of three recurrently connected populations of neurons: $800$ input ($X$) neurons, $800$ excitatory ($E$), and $200$ inhibitory ($I$) neurons. The input neurons do not get any input and generate Poisson spikes independently with a rate of $10 \rm{Hz}$. For the conductance--based integrate-and-fire neuron $i$ in the population $\alpha\in\{E,I\}$ the dynamics of the membrane potential $V_i^\alpha$ are described by the differential equation
\begin{equation}
\begin{split}
& C_m\frac{dV_i^\alpha}{dt} = \\
& -g_L(V_i^\alpha - V_L) + \sum_{\beta\in\{X,E,I\}}I_i^{\alpha,\beta}(t), \text{ if } V_i^\alpha<V_{th},
\end{split}
\end{equation}
where the membrane capacitance is set to $C_m=0.25 \rm{nF}$ and the leak conductance $g_L=16.7\rm{nS}$. The resting potential is $V_L=-70\rm{mV}$ and the firing threshold $V_{th}=-50\rm{mV}$. After each spike the membrane potential was reset to $V_R=-60\rm{mV}$. $E$ and $I$ neurons have a $2$ and $1\rm{ms}$ refractory period, respectively. $I_i^{\alpha\beta}$ is the input current neuron $i$ receives from population $\beta$.

The neurons are connected with probability $p_{connect}=0.2$ and the connections consist of conductance based synapses (for details see the Supplementary Material of Ref.~\cite{renart2010asynchronous}). We draw the conductances for the synapses from a uniform distribution with mean $g^{\alpha\beta}$ and standard deviation $0.5g^{\alpha\beta}$. As in Ref.~\cite{renart2010asynchronous} we set $g^{EE}=2.4\rm{nS}$,$g^{EI}=40\rm{nS}$,$g^{IE}=4.8\rm{nS}$,$g^{II}=40\rm{nS}$ and $g^{EX}=g^{IX}=5.4\rm{nS}$.

For generating data we simulated the network for $T=1000\rm{s}$ and recorded the spike times of a randomly selected subpopulation ($100$ excitatory and $40$ inhibitory neurons). From those, the $30$ excitatory and $10$ inhibitory neurons with the highest firing rates are selected as data for fitting the kinetic Ising model.

To preprocess the data for the Ising model we follow the argument of Ref.~\cite{zeng2013maximum}. The update rate $\gamma$ can be interpreted as the inverse of the width of a neuron's autocorrelation function, which is typically found to be $10\rm{ms}$. Hence we set $\gamma=10^2\rm{Hz}$ and consider a neuron as 'active' for $10\rm{ms}$ after each spike. 
\newpage
\bibliography{bibliography.bib}
\end{document}